\documentclass{article}
\usepackage{ijcai18}

\usepackage{times}
\usepackage{xcolor}
\usepackage{soul}
\usepackage[utf8]{inputenc}
\usepackage[small]{caption}
\usepackage{epsfig}
\usepackage{graphicx}
\usepackage{amsmath}
\usepackage{amssymb}
\usepackage{url}  
\usepackage{graphicx}  
\usepackage{amsmath}
\usepackage{adjustbox}
\usepackage{subfigure}
\usepackage{blindtext}
\usepackage{enumerate}
\usepackage{bm}
\usepackage{dsfont}
\DeclareMathOperator*{\argmin}{min}
\DeclareMathOperator*{\argmax}{max}

\graphicspath{ {figures/} }

\title{Unsupervised Domain Adaptation: from Simulation Engine to the Real World}

\author{Sicheng Zhao, Bichen Wu, Joseph Gonzalez, Sanjit A. Seshia, Kurt Keutzer\\
Department of Electrical Engineering and Computer Sciences, University of California Berkeley\\
\{schzhao,bichen,jegonzal,sseshia,keutzer\}@berkeley.edu}

\begin{document}

\maketitle

\begin{abstract}
Large-scale labeled training datasets have enabled deep neural networks to excel on a wide range of benchmark vision tasks.
However, in many applications it is prohibitively expensive or time-consuming to obtain large quantities of labeled data. To cope with limited labeled training data, many have attempted to directly apply models trained on a large-scale labeled source domain to another sparsely labeled target domain. Unfortunately, direct transfer across domains often performs poorly due to \emph{domain shift} and \emph{dataset bias}.
Domain adaptation is the machine learning paradigm that aims to learn a model from a source domain that can perform well on a different (but related) target domain. In this paper, we summarize and compare the latest unsupervised domain adaptation methods in computer vision applications. We classify the non-deep approaches into sample re-weighting and intermediate subspace transformation categories, while the deep strategy includes discrepancy-based methods, adversarial generative models, adversarial discriminative models and reconstruction-based methods. We also discuss some potential directions.
\end{abstract}

\section{Introduction}
\label{sec:Introduction}

\begin{figure}[!t]
\begin{center}
\subfigure{
\label{fig:Cityscapes}
\includegraphics[height=0.243\linewidth]{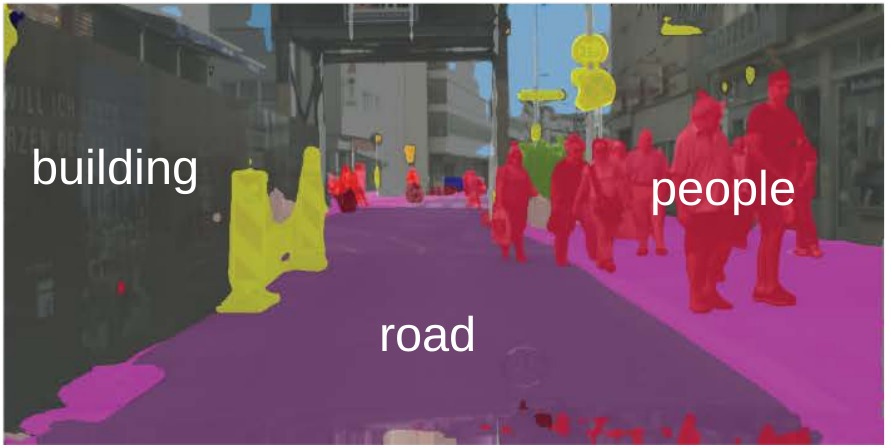}
}
\subfigure{
\label{fig:SanFranciscoDashcam}
\includegraphics[height=0.243\linewidth]{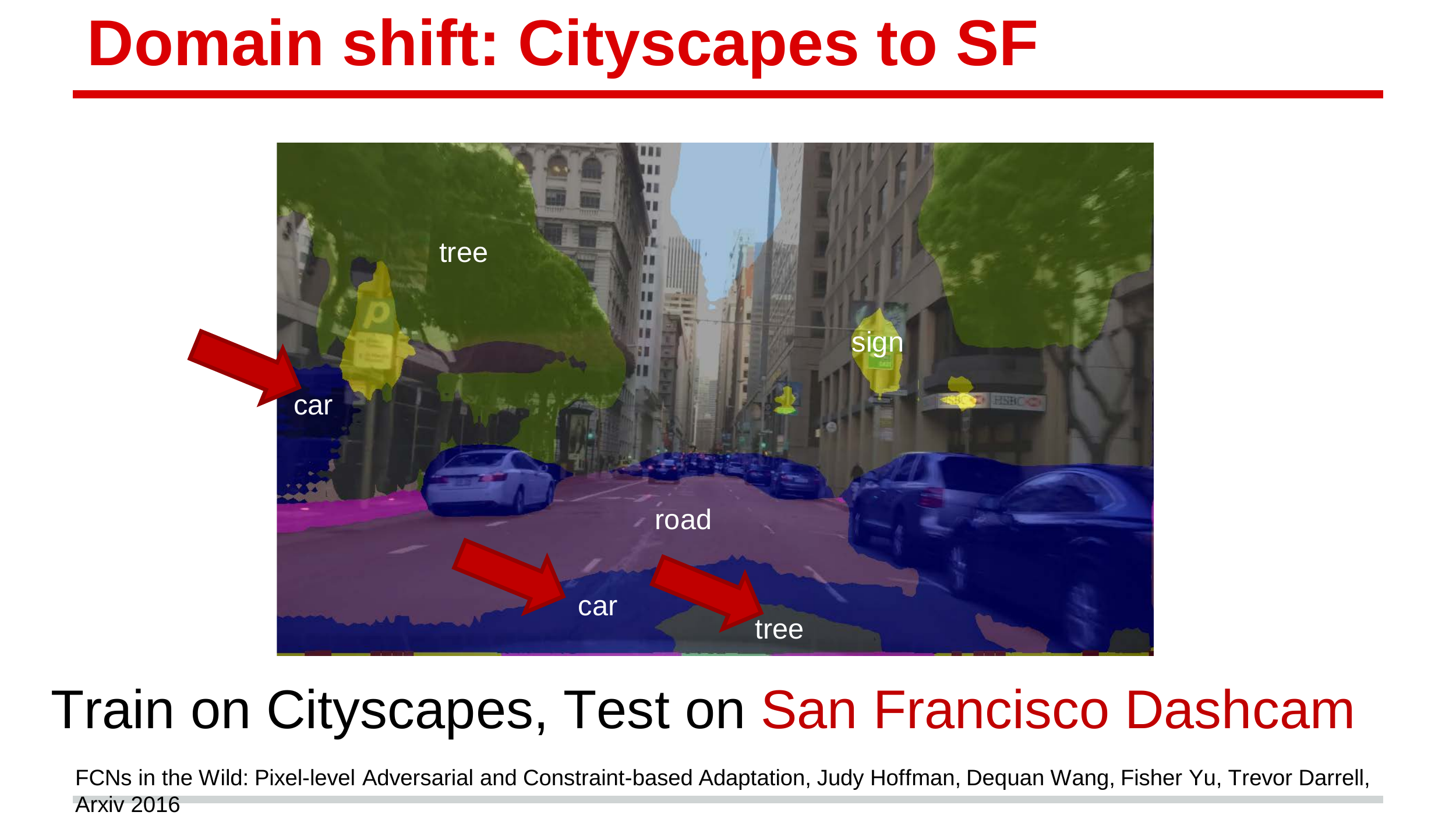}
}
\caption{An example of dataset bias or domain shift~\protect\cite{hoffman2016fcns}. The FCN segmentation model trained on the Cityscapes is tested on the left image from the Cityscapes and the right image from the San Francisco Dashcam. The regions pointed out by red arrows are segmented with incorrect class labels.}
\label{fig:DatasetBias}
\end{center}
\end{figure}

Deep neural networks have achieved satisfying performance in various vision tasks with large-scale labeled training data. For example, the classification error of the ``Classification + localization with provided training data'' task in the Large Scale Visual Recognition Challenge has reduced from 0.28 in 2010 to 0.022 in 2017 (http://image-net.org/challenges/LSVRC/2017), even outperforming humans.
However, in many applications, it is difficult to obtain large amount of labels, as labeling is expensive and time-consuming. Directly generalizing the models trained on one large-scale labeled source domain to another related and unlabeled target domain usually may not perform well (see Figure~\ref{fig:DatasetBias}), because of the \emph{dataset bias}~\cite{torralba2011unbiased} or \emph{domain shift}, i.e. the joint probability distributions of observed data and labels are different in the two domains.

One may argue that we can fine-tune the pretrained models in the target domain.
However, fine-tuning still requires considerable quantities of labeled training data, which may be not available for many applications.
For example, in fine-grained recognition, only experts are able to provide reliable labeled data~\cite{gebru2017fine}; in segmentation, it took about 90 minutes to label each image in the Cityscapes dataset~\cite{cordts2016cityscapes}; in autonomous driving, the substantial traffic data obtained with different sensors, such as 3D LiDAR point clouds, are difficult to label~\cite{wu2017squeezeseg}.


Meanwhile, recent progress in graphics and simulation infrastructure can create large amount of simulated and labeled data. For example, CARLA (http://www.carla.org) and GTA-V (https://www.rockstargames.com/V) are two popular simulators for autonomous driving research.
Several recent efforts \cite{shrivastava2017learning,hoffman2017cycada} have studied models trained on simulated data.
Unfortunately, while models trained on simulated data perform well on simulated data they often do not transfer to real-world settings.
While there are ongoing efforts to make simulations more realistic, it is very difficult to model all the characteristics of real data~\cite{shrivastava2017learning}. Therefore, transferring the labeled data in the simulation domain to the real-world domain is a promising alternative.


Domain adaptation, also known as domain transfer, is a form of transfer learning which aims to learn a model from a source domain that can generalize to a different (but related) target domain.
With increasing demands in different applications, domain adaptation has recently attracted significant interest in artificial intelligence.
In this paper, we survey recent unsupervised domain adaptation approaches with computer vision applications, compare their differences and (dis)advantages, and discuss potential research directions.

\section{Notations and Problem Definition}
\label{sec:ProblemDefinition}

We attempt to introduce a standard definition of the variables and models to enable effective comparisons. Let $\textbf{x}$ and $y$ respectively denote the input data and output label variables, drawn from a specific domain probability distribution $P(\textbf{x},y)$.
In typical domain adaptation, there is one source domain and one target domain.
Suppose the source data and corresponding labels drawn from the source distribution $P_S(\textbf{x},y)$ are $\textbf{X}_S$ and $Y_S$, and the target data and corresponding labels drawn from the target distribution $P_T(\textbf{x},y)$ are $\textbf{X}_T$ and $Y_T$. Corresponding marginal distributions include $P_S(\textbf{x})$, $P_S(y)$, $P_T(\textbf{x})$, $P_T(y)$, and conditional distributions include $P_S(\textbf{x}|y)$, $P_S(y|\textbf{x})$, $P_T(\textbf{x}|y)$, $P_T(y|\textbf{x})$. Two fundamental sources of variation between the two domains are (1) covariate shift, $P_S(y|\textbf{x})=P_T(y|\textbf{x})$ for all $x$, but $P_S(\textbf{x})\neq P_T(\textbf{x})$; (2) concept drift, $P_S(y|\textbf{x})\neq P_T(y|\textbf{x})$. Specifically, the source dataset is $D_S=\{\textbf{X}_S,Y_S\}=\{(\textbf{x}_S^i,y_S^i)\}_{i=1}^{N_S}$, the target dataset is $D_T=\{\textbf{X}_T,Y_T\}=\{(\textbf{x}_T^j,y_T^j)\}_{j=1}^{N_T}$, where $N_S$ and $N_T$ are the number of source samples and target samples, $\textbf{x}_S^i\in \mathds{R}^{d_S}$ and $\textbf{x}_T^j\in \mathds{R}^{d_T}$ are referred as an observation in the source domain and the target domain, and $y_S^i$ and $y_T^j$ are corresponding class labels.
Unless otherwise specified, we assume (1) $d_S=d_T$, which indicates that the data from different domains are observed in the same feature space but exhibit different distributions; (2) $y_S^i\in \mathcal{Y}, y_T^j\in \mathcal{Y}$, where $\mathcal{Y}$ is the class label space. Generally, $Y_S$ is fully labeled and $Y_T$ is unlabeled or partially labeled. Suppose the number of labeled target samples is $N_{TL}$, the domain adaptation problem can be classified into different categories:

(1) \emph{unsupervised domain adaptation}, when $N_{TL}=0$;

(2) \emph{fully supervised domain adaptation}, when $N_{TL}=N_T$;

(3) \emph{semisupervised domain adaptation}, otherwise.

Further, if there is more than one source domain $S_1,S_2,\cdots,S_{N_{MS}}$, the task turns to \emph{multi-source domain adaptation}~\cite{sun2015survey,bhatt2016multi}. If $d_S\neq d_T$, the problem is named as \emph{heterogeneous domain adaptation}~\cite{li2014learning,hubert2016learning}. Please note that multi-source or heterogeneous domain adaptation usually accompanies one of the above three supervision cases.

We focus on the survey of unsupervised domain adaptation (UDA) in one-source and homogeneous settings, i.e. $N_{TL}=0, N_{MS}=1, d_S=d_T$. The goal is to learn a model $f$ with parameter $\bm{\theta}_f$ that can correctly predict a sample from the target domain based on $\{\textbf{X}_S,Y_S\}$ and $\{\textbf{X}_T\}$. We take the multi-class classification task as an example and optimize the loss function from the source domain as
\begin{equation}
\begin{aligned}
&\mathcal{L}_{c}(\textbf{X}_S,Y_S;\bm{\theta}_f,\bm{\theta}_h)=\mathbb{E}_{(\textbf{x}_S,y_S)\sim P_S(\textbf{x},y)}l(\textbf{x}_S,y_S),\\
&l(\textbf{x}_S,y_S)=dis(f(h(\textbf{x}_S;\bm{\theta}_h);\bm{\theta}_f),y_S),
\end{aligned}
\end{equation} 
where $h$ is a feature mapping with parameter $\bm{\theta}_h$, $dis$ is a distance function between the predicted label and the ground truth.
Please note that here $\mathbb{E}_{(\textbf{x}_S,y_S)\sim P_S(\textbf{x},y)}$ ($\mathbb{E}_{(\textbf{x}_S,y_S)\sim P_S}$ for short) is equivalent to $\sum_{i=1}^{N_S}$ by replacing $\textbf{x}_S$ with $\textbf{x}_S^i$ and $y_S$ with $y_S^i$ when computing the empirical loss. For simplicity, we omit the parameters of the functions below.

For semisupervised and fully supervised domain adaptation, please refer to~\cite{patel2015visual} and \cite{tzeng2015simultaneous}, respectively. For other transfer learning paradigms, such as self-taught learning and multi-task learning, please refer to~\cite{pan2010survey}. The survey~\cite{patel2015visual} is mainly about the early methods without much discussion on recent deep learning based methods, and the survey~\cite{csurka2017domain} reviews almost all categories of domain adaptation methods briefly but not computationally.

\section{Non-deep Approaches}
\label{sec:Non-deep}

The early unsupervised domain adaptation (UDA) \mbox{methods} are mainly non-deep approaches, which aims to match the feature distributions between the source domain and the target domain. Roughly, these methods can be divided into two categories: (1) sample re-weighting and (2) intermediate subspace transformation.


\subsection{Sample Re-weighting}
\label{ssec:Re-weighting}

\citeauthor{huang2007correcting}~\shortcite{huang2007correcting} proposed to re-weight the training samples such that the means of the source and target domains in a reproducing kernel Hilbert space (RKHS) $\mathcal{H}$ are close. In this way, the computation of the objective function with \mbox{respect} to (w.r.t.) $P_T$ can be transformed to w.r.t. $P_S$. Based on the covariate shift assmuption, the transformation is simply accomplished with the coefficient $P_T(\textbf{x})/P_S(\textbf{x})$, i.e.,
\begin{equation}
\mathbb{E}_{(\textbf{x}_T,y_T)\sim P_T}l(\textbf{x}_T,y_T)=\mathbb{E}_{(\textbf{x}_T,y_T)\sim P_S}\beta(\textbf{x}_T,y_T)l(\textbf{x}_T,y_T),
\end{equation}
where $\beta(\textbf{x}_T,y_T)=P_T(\textbf{x}_T,y_T)/P_S(\textbf{x}_T,y_T)=P_T(\textbf{x}_T)/P_S(\textbf{x}_T)=\beta(\textbf{x}_T)$. Instead of firstly estimating $P_S$, $P_T$ and subsequently computing $\beta$, \citeauthor{huang2007correcting} designed a more robust and flexible strategy, named kernel mean matching, to infer $\beta$ by
\begin{equation}
\begin{aligned}
\mathop{\argmin}_\beta&\parallel \mathbb{E}_{\textbf{x}_T\sim P_T}h(\textbf{x}_T)-\mathbb{E}_{\textbf{x}_T\sim P_S}\beta(\textbf{x}_T)h(\textbf{x}_T)\parallel,\\
&\text{s.t. }\, \ \beta(\textbf{x}_T)\geq 0 \quad \text{and} \quad \ \mathbb{E}_{\textbf{x}_T\sim P_S}\beta(\textbf{x}_T)=1.
\end{aligned}
\end{equation}

To learn domain-invariant features, \citeauthor{gong2013connecting}~\shortcite{gong2013connecting} exploited the existence of landmarks, which are defined as a subset of labeled samples from the source domain that are distributed similarly to the target domain. Suppose the landmark indicator of the source samples is $\bm{\alpha}=\{\alpha_i\in\{0,1\}\}$, the difference in sample means of the source and target domains in the RKHS is minimized to select landmarks,
\begin{equation}
\mathop{\argmin}_{\bm{\alpha}}\left\Vert \frac{1}{\sum_i \alpha_i}\sum\nolimits_i \alpha_i h(\textbf{x}_S^i)-\frac{1}{N_T}\sum\nolimits_j h(\textbf{x}_T^j)\right\Vert_{\mathcal{H}}^2,
\end{equation}
with balanced label constraints. Based on multi-scale kernel mappings, different sets of landmarks $L_q(q=1,\cdots,N_q)$ are selected. By augmenting the original target domain and weakening the original source domain with landmarks $D_T^q=D_T\bigcup L_q,D_S^q=D_S\setminus L_q$ , a cohort of auxiliary tasks are created, where the distinction across domains are blurred. The solutions of the auxiliary tasks using the geodesic flow kernel algorithm~\cite{gong2012geodesic} form the basis to compose invariant features for the original task. The SVM trained using the concatenation of the invariant features and the labels of $\{L_q\}$ is used to generalize to the target domain.

\begin{table*}[!t]
\centering\scriptsize
\caption{Comparison of different intermediate subspace transformation methods, where `\# sub' indicates the number of subspaces, `linearity' represents the linearity of the subspaces, `NN' is short for nearest neighbor. Euclidean distance is employed for $dis$ in all methods.}
\begin{tabular}
{c | c c c c c c c c}
\hline
& method & \# sub & the object of PCA & linearity & $f$ & $h$ \\
\hline
\cite{gopalan2011domain} & sampling geodesic flow & a set & original samples & linear & NN & SURF\\
\cite{gong2012geodesic} & geodesic flow kernel & a set & original samples & non-linear  & 1-NN & SURF \\
\cite{gopalan2014unsupervised} & mercer kernel & a set & kernel Gram matrix & non-linear & NN & SURF\\
\cite{ni2013subspace} & dictionary learning & a set & original samples & linear & SVM & dictionary\\
\cite{fernando2013unsupervised} & subspace alignment & 2 & original samples & linear & 1-NN, SVM &  SURF, dense SIFT \\
\hline
\end{tabular}
\label{tab:Subspace}
\end{table*}



Both methods use Euclidean distance for $dis$, SVM for $f$, while Gaussian kernel~\cite{huang2007correcting} and geodesic flow kernel~\cite{gong2013connecting} are selected for $h$. The assumption of the former method that $P_S(y|\textbf{x})=P_T(y|\textbf{x})$ and that the support of $P_T$ is contained in the support of $P_S$ is too strong, while the landmarks of the latter method may not exist.


\subsection{Intermediate Subspace Transformation}
\label{ssec:IntermediateSubspace}
\citeauthor{gopalan2011domain}~\shortcite{gopalan2011domain} proposed a sampling geodesic flow (SGF)-based method to exploit low-dimensional structures. SGF models each domain in a $s$-dimensional linear subspace and embeds them to two points on a Grassmann manifold. The collection of all $s$-dimensional subspaces form the Grassmannian $\mathds{G}(s,N)$. They assumed that if the two points are close, the two domains are similar to each other. Let $\textbf{B}_S, \textbf{B}_T\in \mathds{R}^{N\times s}$ denote the basis of the PCA subspaces for the source and target domains, respectively. By viewing $\mathds{G}(s,N)$ as a quotient space, the geodesic path in $\mathds{G}(s,N)$ starting from $\textbf{B}_S$ is given by a one-parameter exponential flow
\begin{equation}
\bm{\Phi}(t)=\textbf{Q}\exp(t\textbf{C})\textbf{Q}^\mathrm{T}\textbf{B}_S, t\in[0,1],
\end{equation}
where $\exp$ is the matrix exponential, $\textbf{Q}\in\mathds{G}(s,N)$ and $\textbf{Q}^\mathrm{T}\textbf{B}_S=\bigg(\begin{matrix}\textbf{I}_s\\ \textbf{0}_{N-s,s}\end{matrix}\bigg)$, $\textbf{C}=\bigg(\begin{matrix}\textbf{0} & \textbf{A}^\mathrm{T}\\-\textbf{A} & \textbf{0}\end{matrix}\bigg)$ is a skew-symmetric, block-diagonal matrix with \textbf{A} specifying the direction and the speed of geodesic flow, and $\textbf{I}_s$ is the identity matrix of size $s$. After obtaining $\textbf{A}$ with inverse exponential mapping, a sequence of intermediate subspaces between $\textbf{B}_S$ and $\textbf{B}_T$ can be obtained by selecting discrete $t$ from 0 to 1.

To address the limitations of SGF, such as sampling strategy selection and parameter tuning, \citeauthor{gong2012geodesic}~\shortcite{gong2012geodesic} extended the GF in a kernel framework. Suppose $\textbf{R}_S\in \mathds{R}^{N\times(N-s)}$ is the orthogonal complement to $\textbf{B}_S$, i.e., $\textbf{R}_S^\mathrm{T}\textbf{B}_S=\textbf{0}$. Using the canonical Euclidean metric for the Riemannian manifold, the GF between $\textbf{B}_S$ and $\textbf{B}_T$ on the manifold is parameterized as a path connecting the two subspaces
\begin{equation}
\bm{\Phi}(t)=\textbf{B}_S\textbf{U}_1\bm{\Gamma}(t)-\textbf{R}_S\textbf{U}_2\bm{\Sigma}(t), t\in[0,1],
\end{equation}
with the constraints $\bm{\Phi}(0)=\textbf{B}_S$ and $\bm{\Phi}(1)=\textbf{B}_T$, $\textbf{U}_1\in\mathds{R}^{s\times s}$ and $\textbf{U}_2\in\mathds{R}^{(N-s)\times s}$ are orthonormal matrices, $\bm{\Gamma}$ and $\bm{\Sigma}$ are diagonal matrices. The GF is viewed as a collection of infinite features varying gradually from the source to the target with the inner products induced by a kernel function.

%
\citeauthor{gopalan2014unsupervised}~\shortcite{gopalan2014unsupervised} proposed a high-dimensional RKHS approach using Mercer kernel mapping to account for non-linear data. Specifically, kernel Gram matrix $\textbf{K}_{S}\in\mathds{R}^{N_S\times N_S}$ is constructed from all the source samples. Corresponding to the top $s$ eigenvalues of $\textbf{K}_{S}$, the eigenvectors constitute a matrix $\textbf{A}_{S}\in\mathds{R}^{N_S\times s}$. Similarly, in the target domain, $\textbf{K}_{T}\in\mathds{R}^{N_T\times N_T}$, $\textbf{A}_{T}\in\mathds{R}^{N_T\times s}$. For any source sample $\textbf{x}_S^i$ and target sample $\textbf{x}_T^j$, the kernel PCA representation turns to
\begin{equation}
\bm{\alpha}_S=\textbf{A}_{S}^\mathrm{T}\textbf{k}_S(\textbf{x}_S^i),\bm{\alpha}_T=\textbf{A}_{T}^\mathrm{T}\textbf{k}_T(\textbf{x}_T^j),
\end{equation}
which consists of the projected coefficients of the original sample onto the orthonormal principal components in the RKHS, $\textbf{k}_S$ and $\textbf{k}_T$ are $N_S\times 1$ and $N_T\times 1$ vectors. By KPCA, the samples of each domain are represented as uncorrelated and Gaussian distributed. With the incremental change of orthonormal principal components from the source to the target, the projected coefficients of an original sample simply incrementally scale from the Gaussian variances of the source to those of the target.

Instead of using different kernels, \citeauthor{ni2013subspace}~\shortcite{ni2013subspace} proposed to interpolate subspaces through dictionary learning. $N_\textbf{D}$ intermediate domain dictionaries $\{\textbf{D}_k\in\mathds{R}^{N\times N_d}\}_{k=0}^{N_{\textbf{D}}}$ are learned to gradually adapt the source to the target. $\textbf{D}_0$ is the dictionary learned from $\textbf{X}_S$ using standard dictionary learning methods and $N_d$ is the number of atoms in the dictionary. Specifically, the domain dictionary $\textbf{D}_{k+1}$ is learned by estimating $\Delta\textbf{D}_{k}$ from its coherence with $\textbf{D}_{k},k\in[0,N_\textbf{D}-1]$ and the reconstruction residue $\textbf{J}_{k}$ of the target data
\begin{equation}
\begin{aligned}
\mathop{\argmin}_{\Delta\textbf{D}_{k}}&\parallel\textbf{J}_{k}-\Delta\textbf{D}_{k}\bm{\Gamma}_k\parallel^2_F+\lambda \parallel\Delta\textbf{D}_{k}\parallel^2_F,\\
&\textbf{J}_{k}=\parallel \textbf{X}_{T}-\textbf{D}_{k}\bm{\Gamma}_k\parallel^2_F,
\end{aligned}
\end{equation}
where $\bm{\Gamma}_k\in\mathds{R}^{N_d\times N_T}$ is the sparse coefficients of decomposed with $\textbf{D}_{k}$ and $\parallel .\parallel^2_F$ is the Frobenius norm. The final dictionary $\textbf{D}_{N_\textbf{D}}$ that best represents the target data in terms of reconstruction error is taken as the target domain dictionary.

The above methods need to build a set of intermediate subspaces. \citeauthor{fernando2013unsupervised}~\shortcite{fernando2013unsupervised} proposed to project each source and target sample to its respective subspace and learn a linear transformation to align the source subspace to the target one. Concretely, to align subspace $\textbf{B}_{S}$ to $\textbf{B}_{T}$, a transformation matrix $\textbf{M}$ is learned by minimizing the Bregman matrix divergence
\begin{equation}
\mathop{\argmin}_{\textbf{M}}\parallel \textbf{B}_{S}\textbf{M}-\textbf{B}_{T}\parallel^2_F=\mathop{\argmin}_{\textbf{M}}\parallel \textbf{M}-\textbf{B}_{S}^{\prime}\textbf{B}_{T}\parallel^2_F,
\end{equation}
where $\textbf{B}_{S}^{\prime}$ is orthonormal, i.e. $\textbf{B}_{S}^{\prime}\textbf{B}_{S}=\textbf{I}_s$.

The summarization and comparison of these methods is illustrated in Table~\ref{tab:Subspace}. Please note that $h$ and $f$ here can be easily generalized to other features and classifiers.


\section{Deep Unsupervised Domain Adaptation}
\label{sec:Deep}
With the advent of deep learning, emphasis has been shifted to learning domain invariant features in an end-to-end fashion.
Typically, a conjoined architecture with two streams is employed to represent the models for the source and target domains, respectively~\cite{zhuo2017deep}. Besides the traditional classification loss based on the labeled source data, deep UDA models are usually trained jointly with another loss to deal with the domain shift, such as discrepancy loss, adversarial loss and reconstruction loss. We divide these methods into four categories based on the domain shift loss and generative/discriminative settings.

Let $\mathds{1}$ and $\sigma$ respectively denote the indicator function and the softmax function. Typically, the cross-entropy loss is employed as the classification loss, i.e.,
\begin{equation}
l(\textbf{x}_S,y_S)=\sum\nolimits_{k=1}^K\mathds{1}_{[k=y_S]}\log(\sigma(f(h(\textbf{x}_S)))).
\end{equation}
The loss functions of the deep methods discussed below is the joint combination of the cross-entropy loss and another new-designed loss. Unless otherwise specified, we will discuss the new-designed loss only in the following subsections.

\subsection{Discrepancy-based Methods}
\label{ssec:Discrepancy}

Discrepancy-based methods explicitly measure the discrepancy between the source and target domains on corresponding activation layers of the two network streams. \citeauthor{long2015learning} \shortcite{long2015learning} designed a Deep Adaptation Network (DAN), where the discrepancy is defined as the sum of the multiple kernel variant of maximum mean discrepancies (MK-MMD) between the fully connected (FL) layers
\begin{equation}
\mathcal{L}_{MK}=\sum_{l=l_1}^{l_2}\parallel \mathbb{E}_{\textbf{r}_S\sim \textbf{R}_S^l}\phi(\textbf{r}_S)-\mathbb{E}_{\textbf{r}_T\sim \textbf{R}_T^l}\phi(\textbf{r}_T)\parallel_{\mathcal{H}_k}^2,
\label{equ:MKMMD}
\end{equation}
where $l_1$ and $l_2$ are layer indices between which MK-MMD is effective, $\textbf{R}_S^l,\textbf{R}_T^l$ are the $l$th layer hidden representations (embeddings) for the source and target examples. The characteristic kernel $k$ associated with the feature map $\phi$ in RKHS $\mathcal{H}_k$ is $k(\textbf{r}_S,\textbf{r}_T)=<\phi(\textbf{r}_S),\phi(\textbf{r}_T)>$, and is implemented as a linear combination of several positive semi-definite kernels.

\citeauthor{sun2017correlation}~\shortcite{sun2017correlation} proposed correlation alignment (CORAL) to minimize domain shift by aligning the second-order statistics of source and target features of the last FL layer
\begin{equation}
\mathcal{L}_{CORAL}=\parallel \textbf{C}_{S}-\textbf{C}_{T}\parallel_{F}^2/(4N_{FL}^2),
\end{equation}
where $\textbf{C}_{S}$ ($\textbf{C}_{T}$) are the feature covariance matrices
\begin{equation}
\textbf{C}_{S}=(\textbf{R}_{S}^\mathrm{T}\textbf{R}_{S}-(\textbf{1}\textbf{R}_{S}/N_S)^\mathrm{T}(\textbf{1}\textbf{R}_{S}))/(N_S-1),
\end{equation}
where $\textbf{R}_{S}^{ij}$ indicates the $j$th dimension (totally $N_{FL}$ dimension) of the $i$th source feature, $\textbf{1}$ is a column vector with all elements equal to 1. By replacing $N_S$ with $N_T$, $\textbf{R}_{S}^{ij}$ with $\textbf{R}_{T}^{ij}$, we can obtain $\textbf{C}_{T}$.

Apart from the CORAL loss on the last FL layer, \citeauthor{zhuo2017deep}~\shortcite{zhuo2017deep} also incorporated the CORAL loss on the last convolutional (conv) layer. To deal with the high dimension of convolutional layer activations, activation-based attention mapping is employed to distill it into low dimensional representations. Given an activation tensor $\textbf{AT}\in\mathds{R}^{C\times H\times W}$, a mapping function $F_{att}$ that takes $\textbf{AT}$ as input and outputs a spatial attention map is defined as
\begin{equation}
(F_{att}(\textbf{AT})_{i,j})=\sum\nolimits_{ch=1}^{C}|\textbf{AT}_{ch,i,j}|^p.
\label{equ:Attention}
\end{equation}
After transforming $F_{att}$ into vectorized form and applying a logarithmic function, $\log(vec(F_{att}(.)))$ is used to compute the  CORAL loss. The CORAL losses on both the last convolutional layer and the last FC layer are combined.

\begin{table}[!t]
\centering\scriptsize
\caption{Comparison of different discrepancy-based methods, where `loss' indicates the loss objectives without the common cross-entropy loss, `layer' represents the layers that the loss functions on, `weight' indicates whether the weights of the two networks are shared or not, `base net' is the existing network that the compared methods are based on.}
\begin{tabular}
{c | c c c c c c c c}
\hline
& loss & layer & weight & base net \\
\hline
\citeauthor{long2015learning} & MK-MMD & FL & shared & AlexNet \\
\citeauthor{sun2017correlation} & CORAL & last FL & shared & AlexNet\\
\citeauthor{zhuo2017deep} & CORAL & last (conv, FL) & shared & AlexNet\\
\citeauthor{rozantsev2016beyond} & weight, MMD & all & linear & AlexNet, LeNet\\
\hline
\end{tabular}
\label{tab:Discrepancy}
\end{table}

The above methods all adopt sharing weights of the two streams of the Siamese architecture. On the contrary, \citeauthor{rozantsev2016beyond}~\shortcite{rozantsev2016beyond} relaxed the sharing weight constraint by assuming that the weights of corresponding layers in the two models remain linearly related. Besides the standard classification loss, another two regularizers are jointly optimized. One is weight regularizer $\mathcal{L}_W$, representing the loss between corresponding layers of the two streams
\begin{equation}
\mathcal{L}_W=\sum\nolimits_{l}\exp(\parallel a_l\bm{\theta}_{f_S}^l+b_l-\bm{\theta}_{f_T}^l\parallel^2)-1,
\end{equation}
where $\bm{\theta}_{f_S}^l$ and $\bm{\theta}_{f_T}^l$ are the parameters of the $l$th layer of the source and target streams, $a_l$ and $b_l$ are scalar parameters that are different across layers. The other is the unsupervised regularizer $\mathcal{L}_{MMD}$, encoding the MMD measure and favoring similar distributions of the source and target representations. $\mathcal{L}_{MMD}$ is of the form Equ.~(\ref{equ:MKMMD}), except that in implementation there is only one kernel.

The comparison of these methods is summarized in Table~\ref{tab:Discrepancy}. Since the domain invariant features of deep UDA methods are learned end-to-end, we will not compare the detailed difference on $dis$, $f$ and $h$ unless otherwise specified.

\subsection{Adversarial Generative Models}
\label{ssec:Generative}

Adversarial generative models combine the domain discriminative model with a generative component generally based on generative adversarial nets (GANs)~\cite{goodfellow2014generative}, which includes a generator $g$ with parameter $\bm{\theta}_g$ and a discriminator $d$ with parameter $\bm{\theta}_d$. $g$ takes random noise $\textbf{z}$ as input to generate a virtual image, and $d$ takes the output of $g$ and real images $\textbf{x}$ as input to classify whether an image is real or generated. The learning process is that $d$ tries to maximize the probability of correctly classifying real images and generated images, while $g$ tries to generate images to maximize the probability of $d$ making a mistake. In other words, the following two-player minimax game is played
\begin{equation}
\begin{aligned}
\mathop{\argmin}_{g}\mathop{\argmax}_{d}\mathcal{L}_{G}(d,g)&=\mathbb{E}_{\textbf{x}\sim P_{\textbf{x}}}\log d(\textbf{x})\\
&+\mathbb{E}_{\textbf{z}\sim P_{\textbf{z}}}\log (1-d(g(\textbf{z}))).\\
\end{aligned}
\label{equ:GAN}
\end{equation}
When optimizing $g$, the loss $\mathcal{L}_{g}$ only includes the second part. While optimizing $d$, the loss $\mathcal{L}_{d}$ includes both.

\begin{table*}[!t]
\centering\scriptsize
\caption{Comparison of different adversarial generative models, where `loss' indicates the loss objectives without the common cross-entropy loss, `layer' represents the layers that the loss functions on, `weight' indicates whether the weights of different GANs are shared or not, the number after `ResNet-' is the number of ResNet blocks.}
\begin{tabular}
{c | c c c c c c c c}
\hline
& loss & input of GAN & weight & $g$ base net & $d$ base net & $f$ base net \\
\hline
\cite{liu2016coupled} & Coupled GAN & $\textbf{z}$ & partially shared & self-defined  & LeNet & self-defined\\
\cite{shrivastava2017learning} & GAN with new $\mathcal{L}_{g}$ & $\textbf{x}_S$ & - & ResNet-4 & self-defined & self-defined\\
\cite{bousmalis2017unsupervised} & GAN, masked-PMSE & $\textbf{z},\textbf{x}_S$ & - & ResNet-3 & self-defined & AlexNet\\
\cite{hoffman2017cycada} & CycleGAN, semantic, feature & $\textbf{x}_S,\textbf{x}_T, fea$ & unshared & ResNet-7 & self-defined &LeNet, VGG16, DRN \\
\cite{kang2018deep} & CycleGAN, attention map & $\textbf{x}_S,\textbf{x}_T$ & unshared & ResNet-50 & self-defined & ResNet, AlexNet\\
\hline
\end{tabular}
\label{tab:Generative}
\end{table*}

The Coupled Generative Adversarial Networks (CoGAN)~\cite{liu2016coupled} is composed of a tuple of GANs, each corresponding to one domain. CoGAN can learn a joint distribution of multi-domain images without existence of corresponding images in different domains, simply by enforcing a weight-sharing constraint to the layers that are responsible for decoding high-level semantics. For example, in the unsupervised domain adaptation situation discussed in this paper, CoGAN consists of a pair of GANs, each is responsible for synthesizing images in one domain (source or target). In such cases, the CoGAN corresponds a constrained minimax game of two teams, each with two players
\begin{equation}
\begin{aligned}
&\mathcal{L}_{G}(d_S,g_S,d_T,g_T)=\mathbf{E}_{\textbf{x}_S\sim P_S}\log d_S(\textbf{x}_S)\\
&+\mathbf{E}_{\textbf{z}\sim P_\textbf{z}}\log (1-d_S(g_S(\textbf{z})))+\mathbf{E}_{\textbf{x}_T\sim P_T}\log d_T(\textbf{x}_T)\\
&+\mathbf{E}_{\textbf{z}\sim P_\textbf{z}}\log (1-d_T(g_T(\textbf{z}))),
\end{aligned}
\end{equation}
where $g_S$ ($g_T$) and $d_S$ ($d_T$) are the generator and discriminator of the source (target) GAN, $\bm{\theta}_{g_S^{(i)}}=\bm{\theta}_{g_T^{(i)}}, i=1,2,\cdots,m$ and $\bm{\theta}_{d_S^{(n_S-j)}}=\bm{\theta}_{d_T^{(n_T-j)}}, j=0,1,2,\cdots,n-1$, which indicate the weight-sharing constraint of the first $m$ layers for the generator and the last $n$ layers for the discriminator.

\citeauthor{shrivastava2017learning}~\shortcite{shrivastava2017learning} proposed simulated and unsupervised learning (SimGAN) to improve the realism of a simulator's output using unlabeled real data. The discriminator's loss in SimGAN is the same as that of traditional GAN, while a self-regularization loss is added in the refiner (generator) loss to ensure that the refined data do not change much, which aims to preserve the annotation information
\begin{equation}
\mathcal{L}_{g}=\mathbb{E}_{\textbf{x}_S\sim P_{S}}[\log (1-d(g(\textbf{x}_S)))+\lambda \parallel g(\textbf{x}_S)-\textbf{x}_S\parallel_1].
\end{equation}
Another two improvements in SimGAN are that the discriminator's output is a multiple dimensional probability map of patches to reflect the receptive field and that the discriminator is trained using a history of refined images rather than only the ones from the current refiner network, which aims to stabilize training.

\citeauthor{bousmalis2017unsupervised}~\shortcite{bousmalis2017unsupervised} also exploited GANs to \mbox{adapt} source images to appear as if they are drawn from the target domain. The generator in this model is conditioned on both a noise vector and an image from the source domain. By decoupling the process of domain adaptation from the task-specific architecture, the model can generalize to object classes unseen during the training phase. Furthermore, to penalize large low-level differences between source and generated images for foreground pixels only, the model learns to minimize a masked Pairwise Mean Squared Error (PMSE) which only calculates the masked pixels (foreground) of the source and the generated images. The joint objective is
\begin{equation}
\mathop{\argmin}_{\bm{\theta}_g,\bm{\theta}_f}\mathop{\argmax}_{\bm{\theta}_d}\alpha\mathcal{L}_d(d,g)+\beta\mathcal{L}_c(f,g)+\gamma\mathcal{L}_e(g).
\end{equation}
Similar to Equ.~(\ref{equ:GAN}), $\mathcal{L}_d$ represents the domain loss by replacing $\textbf{x}$ with $\textbf{x}_T$ and $\textbf{z}$ with $\textbf{x}_S,\textbf{z}$. $\mathcal{L}_c$ is the cross-entropy loss. $\mathcal{L}_e$ is the masked-PMSE loss
\begin{equation}
\begin{aligned}
\mathcal{L}_e(g)=&\mathbb{E}_{\textbf{x}_S\sim P_{S},\textbf{z}\sim P_{\textbf{z}}}[\parallel (\textbf{x}_S-g(\textbf{x}_S,\textbf{z}))\circ\textbf{m}\parallel_2^2/N_{\textbf{x}_S}\\
&-((\textbf{x}_S-g(\textbf{x}_S,\textbf{z}))\textbf{m})^2/N_{\textbf{x}_S}^2],
\end{aligned}
\end{equation}
where $\textbf{m}\in\mathds{R}^{N_{\textbf{x}_S}}$ is a binary mask, $N_{\textbf{x}_S}$ is the number of pixels in input $\textbf{x}_S$, and $\circ$ is the Hadamard product.

Based on the cycle-consistency constraints of the CycleGAN~\cite{zhu2017unpaired}, \citeauthor{hoffman2017cycada}~\shortcite{hoffman2017cycada} proposed discriminatively-trained Cycle-Consistent Adversarial Domain
Adaptation (CyCADA), which adapts representations at both the pixel-level and feature-level, enforces cycle-consistency, and leverages a task loss, without the requirement of aligned pairs. A source model $f_S$ is first learned with the cross-entropy loss $\mathcal{L}_{c}(f_S,\textbf{X}_S,\textbf{Y}_S)$. Besides the traditional GAN loss $\mathcal{L}_{G}(g_{ST},d_T,\textbf{X}_T,\textbf{X}_S)$, $g_{ST}$ indicates the generator from source to target, $d_T$ is the corresponding discriminator, there are some other losses. The first is the cross-entropy loss $\mathcal{L}_{c}(f_T,g_{ST}(\textbf{X}_S),\textbf{X}_T,\textbf{Y}_S)$ for the target model $f_T$ based on the translated source image and corresponding labels. Another mapping from target to source is trained to preserve the structure or content of the original sample $\textbf{x}_S$ with the GAN loss $\mathcal{L}_{G}(g_{TS},d_S,\textbf{X}_S,\textbf{X}_T)$. The cycle-consistency is enforced to ensure that mapping a source sample from source to target and back to the source reproduces the original sample by imposing an $L1$ penalty on the reconstruction error
\begin{equation}
\begin{aligned}
&\mathcal{L}_{cyc}(g_{ST},g_{TS},\textbf{X}_S,\textbf{X}_T)=\mathbb{E}_{\textbf{x}_S\sim P_S}\parallel g_{TS}(g_{ST}(\textbf{x}_S))\\
&-\textbf{x}_S\parallel_1+\mathbb{E}_{\textbf{x}_T\sim P_T}\parallel G_{S\rightarrow T}(G_{T\rightarrow S}(\textbf{x}_T))-\textbf{x}_T\parallel_1.
\end{aligned}
\end{equation}
Suppose the predicted label from classifier $f$ is $p(f,\textbf{x})=\arg\argmax(f(\textbf{x}))$, the high semantic consistency is added before and after image translation
\begin{equation}
\begin{aligned}
&\mathcal{L}_{sem}(g_{ST},g_{TS},\textbf{X}_S,\textbf{X}_T,f_S)=\mathcal{L}_{c}(f_S,g_{TS}(\textbf{X}_T),\\
&p(f_S,\textbf{X}_T))+\mathcal{L}_{c}(f_S,g_{ST}(\textbf{X}_S),p(f_S,\textbf{X}_S)).
\end{aligned}
\end{equation}
A feature-level GAN loss $\mathcal{L}_{G}(f_T,d_{feat},f_S(g_{ST}(\textbf{X}_S)),\textbf{X}_T)$ is considered to discriminate between the features or semantics from two image sets as viewed under a task network.
The objective is the joint combination of the above losses.

Besides $\mathcal{L}_{G}(g_{ST},d_T,\textbf{X}_T,\textbf{X}_S)$, $\mathcal{L}_{G}(g_{TS},d_S,\textbf{X}_S,\textbf{X}_T)$ and $\mathcal{L}_{cyc}(g_{ST},g_{TS},\textbf{X}_S,\textbf{X}_T)$, \citeauthor{kang2018deep}~\shortcite{kang2018deep} proposed to impose the attention alignment penalty to reduce the discrepancy of attention maps across domains. The attention map is defined as in Equ.~(\ref{equ:Attention}). The distance betwen the vectorized attention maps of the source and the target networks is penalized to minimize the discrepancy. To make the attention mechanism invariant to the domain shift, the target network is trained with a mixture of real and synthetic data from both
source and target domains.

The comparison of these models is summarized in Table~\ref{tab:Generative}.

\subsection{Adversarial Discriminative Models}
\label{ssec:Discriminative}

Adversarial discriminative models usually employ an adversarial objective with respect to a domain discriminator to encourage domain confusion. Suppose $m_S$ and $m_T$ are the representation mappings of the source and target domains, $d$ is a domain discriminator, which classifies whether a data point is drawn from the source or the target domain. All adversarial losses train the adversarial discriminator using a standard classification loss, typically GAN loss,
\begin{equation}
\begin{aligned}
&\mathop{\argmax}_{d}\mathcal{L}_{a_d}(\textbf{X}_S,\textbf{X}_T,m_S,m_T)=\mathbb{E}_{\textbf{x}_S\sim P_S}\\
&\log d(m_S(\textbf{x}_S))+\mathbb{E}_{\textbf{x}_T\sim P_T}\log (1-d(m_T(\textbf{x}_T))).\\
\end{aligned}
\end{equation}

The loss used to train representation mapping, $\mathcal{L}_{a_m}$, is different in existing methods. The Domain-Adversarial Neural Networks (DANN)~\cite{ganin2016domain} optimizes the mapping to minimize the discriminator loss directly $\mathcal{L}_{a_m}=-\mathcal{L}_{a_d}$, which might be problematic, since early on during training the discriminator converges quickly, causing the gradient to vanish. \citeauthor{tzeng2017adversarial}~\shortcite{tzeng2017adversarial} proposed to use an inverted label GAN loss rather than directly using the minimax loss to split the optimization process into two independent objectives for generator and discriminator in Adversarial Discriminative Domain
Adaptation (ADDA)
\begin{equation}
\mathop{\argmax}_{m_T}\mathcal{L}_{a_m}(\textbf{X}_S,\textbf{X}_T,d)=\mathbb{E}_{\textbf{x}_T\sim P_T}\log d(m_T(\textbf{x}_T)).
\end{equation}

\subsection{Reconstruction-based Methods}
\label{ssec:Reconstruction}

Reconstruction based methods incorporate a reconstruction loss to minimize the difference between the input and the reconstructed
input. \citeauthor{ghifary2015domain}~\shortcite{ghifary2015domain} designed a three-layer Multi-task Autoencoder (MTAE) architecture, which is an autoencoder with multiple (2 when $N_{MS}=1$) output layers, each corresponding to one domain. In MTAE, the input-hidden and hidden-output weights represent shared and domain-specific parameters, respectively. The category-level correspondence across domains is required, which can be implemented by a random selection procedure. Suppose the selected data are $\textbf{X}_S^\prime,\textbf{X}_T^\prime\in\mathds{R}^{N\times N^\prime}$. Let $\overline{\textbf{X}}=[\textbf{X}_S^\prime;\textbf{X}_T^\prime]$, $\overline{\textbf{X}}^l=[\textbf{X}_l^\prime;\textbf{X}_l^\prime]$, $l\in\{S,T\}$, $\overline{\textbf{x}}^\mathrm{T}_i$ and $\overline{\textbf{x}}^{l\mathrm{T}}_i$ be the $i$th row of $\overline{\textbf{X}}$ and $\overline{\textbf{X}}^l$. The feed-forward MTAE reconstruction is
\begin{equation}
\textbf{h}_i=\sigma_{enc}(\textbf{W}^\mathrm{T}\overline{\textbf{x}}_i),f_{\bm{\Theta}^{(l)}}(\overline{\textbf{x}}_i)=\sigma_{dec}(\textbf{V}^{(l)\mathrm{T}}\textbf{h}_i),
\end{equation}
where $\bm{\Theta}^{(l)}=\{\textbf{W},\textbf{V}^{(l)}\}$ contains the matrices of shared and individual weights, $\sigma_{enc}$ and $\sigma_{dec}$ are element-wise non-linear activation functions. Self-domain and between-domain reconstruction tasks are performed truing MTAE Training, which corresponds to minimizing the following objective
\begin{equation}
\sum_{l}\sum_{i=1}^{2N^\prime}\parallel f_{\bm{\Theta}^{(l)}}(\overline{\textbf{x}}_i)-\overline{\textbf{x}}^{l}\parallel_2^2
+\eta (\parallel\textbf{W}\parallel_2^2+\parallel\textbf{V}^{(l)}\parallel_2^2).
\end{equation}

Another representative work is Deep Reconstruction Classification Network (DRCN)~\cite{ghifary2016deep}, which combines a traditional convolutional supervised network for source label prediction with a de-convolutional unsupervised network for target data reconstruction, by viewing the reconstruction network as an approximate of the ideal discriminative representation. The feature mapping parameters of the two streams are shared, while the feature labeling parameters of the supervised network and the feature decoding parameters of the unsupervised network for the reconstruction are learned individually. The reconstruction loss is defined as
\begin{equation}
\mathcal{L}_R=\mathbb{E}_{\textbf{x}_T\sim P_T}\parallel f_R(\textbf{x}_T)-\textbf{x}_T\parallel_2^2,
\end{equation}
where $f_R(\textbf{x}_T)$ is the output of the reconstruction network.

The former MTAE method requires that the number of samples of corresponding category in the two domains should be the same. After sample selection procedure, some important information may be missing. Further, the output of the algorithm is learned features, based on which a classifier (multi-class SVM with linear kernel in this paper) needs to be trained. The latter DRCN method employs an end-to-end strategy, without the requirement of aligned pairs.

\section{Conclusion and Future Directions}
\label{sec:Conclusion}
This paper attempted to provide an overview of recent developments in unsupervised domain adaptation of both non-deep and deep scenarios. Obviously, it cannot cover all the literature on UDA, since too many works have been published recently, and we focused on a representative subset of the latest methods. We summarized these methods with unified variables and formulations, and compared the differences and (dis)advantages. We hope that this survey can help the interested researchers understand UDA better.

We believe that (unsupervised) domain adaptation will continue to be an active and promising research area with broad potential applications, such as autonomous driving. For further studies, researchers can pursue either the methodology or applications of domain adaptation. From the methodology's perspective, incorporating prior knowledge into the adaptation process may lead to performance increase and imaging understanding, since domain shifts are usually caused from the imaging process, such as illumination changes, sensor changes, and viewpoint changes~\cite{patel2015visual}. For adversarial methods, imposing multi-level constraints jointly in the adaptation, such as low-level appearances, mid-level features and high-level semantics, can better preserve the structure and attributes of the source data. In addition, if there are just a few examples for some categories in the labeled source data, how to adapt well in such cases is another challenge. Designing an effective and direct metric to evaluate the quality of adaptation, instead of testing the performance on the target domain, would accelerate the training process of GANs.

For applications, current methods mainly focused on 2D images from the source domain to the target domain. Adapting 3D images, 2D videos, 3D videos or multi-modal data is more challenging and worth studying. Effectively exploring the temporal correlation of videos and the spatial information of 3D data may significantly improve the performance of domain adaptation. For example, adapting the 3D LiDAR point cloud data from synthetic GTA-V to realistic KITTI~\cite{geiger2012we} is very interesting. Meanwhile, existing adaptation methods mainly work on the objective task, such as \mbox{object} classification and scene segmentation, while the adaptation on subjective attributes, such as aesthetics and emotions, has been rarely explored.

\scriptsize\bibliographystyle{named}
\bibliography{DA_Survey}

\end{document}